\documentclass[10pt,twocolumn,letterpaper]{article}

\usepackage[pagenumbers]{cvpr} 
\usepackage{cvpr}              

\usepackage{graphicx}
\usepackage{amsmath}
\usepackage{amssymb}
\usepackage{booktabs}
\usepackage{multirow}
\usepackage[title]{appendix}

\usepackage[pagebackref,breaklinks,colorlinks]{hyperref}

\usepackage[capitalize]{cleveref}
\crefname{section}{Sec.}{Secs.}
\Crefname{section}{Section}{Sections}
\Crefname{table}{Table}{Tables}
\crefname{table}{Tab.}{Tabs.}

\begin{document}

\title{MotionZero:Exploiting Motion Priors for Zero-shot Text-to-Video Generation}

\author{Sitong Su\textsuperscript{1}\hspace{-0.2cm}\and
Litao Guo\textsuperscript{1}\hspace{-0.2cm}\and
Lianli Gao\textsuperscript{1}\hspace{-0.2cm}\and
Heng Tao Shen\textsuperscript{1}\hspace{-0.2cm}\and
Jingkuan Song\textsuperscript{1}\thanks{Jingkuan Song is the corresponding author.} \and 
\textsuperscript{1} University of Electronic Science and Technology of China (UESTC)
\\
{\tt\small {sitongsu9796@gmail.com}}
}

\twocolumn[{%
 \renewcommand\twocolumn[1][]{#1}%
 \maketitle
 \begin{center}
  \centering
  \includegraphics[width=0.98\linewidth]{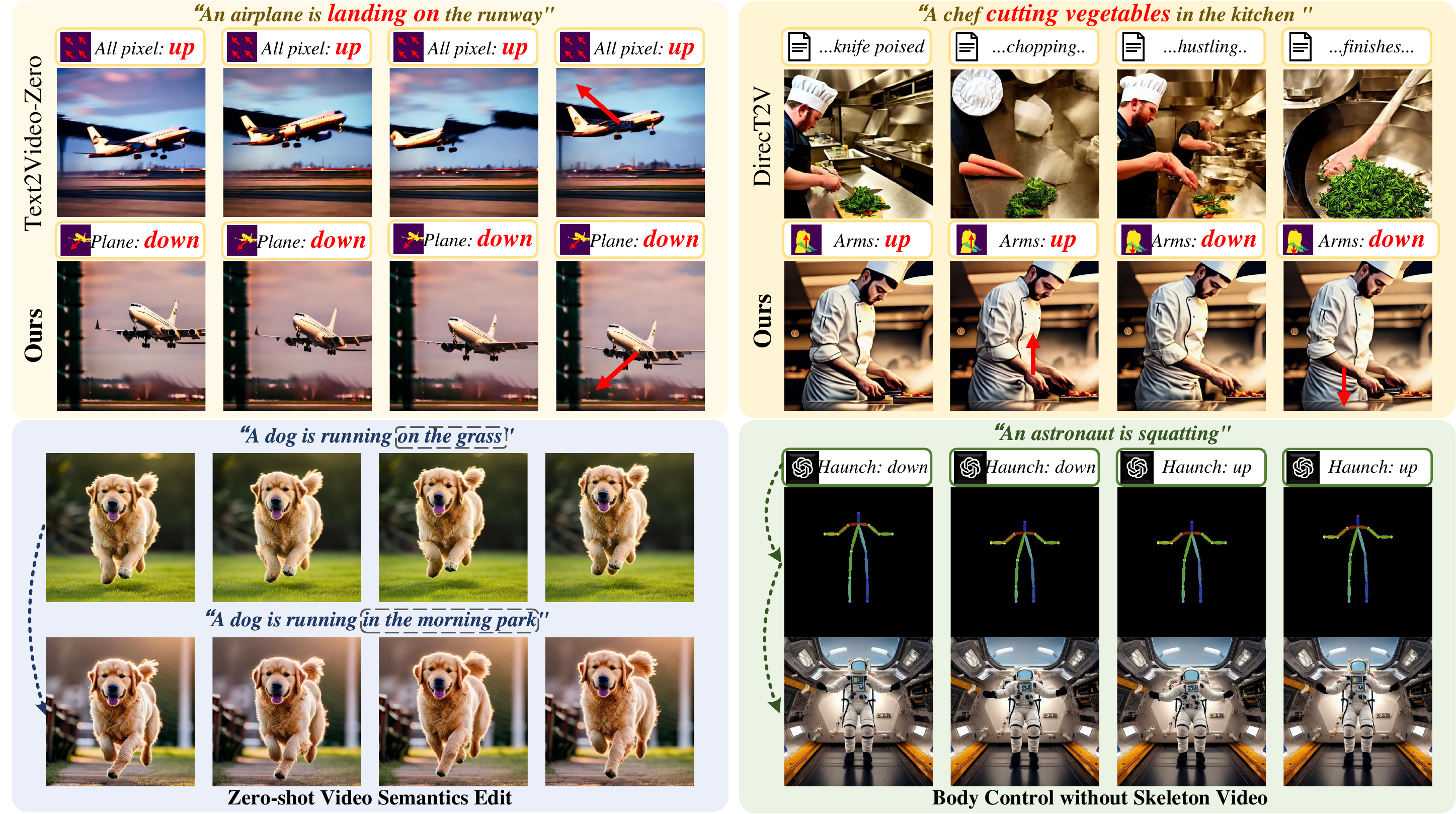}
  \captionof{figure}{\textbf{Row 1-2:} Visual results comparison on Zero-shot Text-to-Video synthesis with two baselines Text-to-Video Zero~\cite{zero} and DirecT2V~\cite{direct2v}. \textbf{Left down in row 3-4:} Our method supports the application of Zero-shot Video Semantics Edit. \textbf{Right down in row 3-4:} Our method supports precise human body control by manipulating skeletons without skeleton videos.}
  \label{frontpage}
 \end{center}%
}]

\begin{abstract}
Zero-shot Text-to-Video synthesis generates videos based on prompts without any videos.
Without motion information from videos, motion priors implied in prompts are vital guidance. For example, the prompt ``airplane landing on the runway" indicates motion priors that the ``airplane" moves downwards while the ``runway" stays static.
Whereas the motion priors are not fully exploited in previous approaches, thus leading to \textbf{two nontrivial issues}:
\textbf{1) }the motion variation pattern remains unaltered and prompt-agnostic for disregarding motion priors;
\textbf{2)} the motion control of different objects is inaccurate and entangled without considering the independent motion priors of different objects.
To tackle the two issues, we propose a prompt-adaptive and disentangled motion control strategy coined as \textbf{MotionZero}, which derives motion priors from prompts of different objects by Large-Language-Models and accordingly applies motion control of different objects to corresponding regions in disentanglement. 
Furthermore, to facilitate videos with varying degrees of motion amplitude, we propose a Motion-Aware Attention scheme which adjusts attention among frames by motion amplitude. 
Extensive experiments demonstrate that our strategy could correctly control motion of different objects and support versatile applications including zero-shot video edit.
\end{abstract}

\section{Introduction}
\label{sec:intro}
Text-to-Video (T2V) refers to synthesizing coherent and textual-aligned videos according to text prompts. Extensive research efforts have been dedicated to T2V because of its vital role in AIGC especially for temporal content creation.

Recently, mainstream methods in T2V are trained on large-scale corpora~\cite{VideoFactory, makeavideo, cogvideo, wu2021godiva, Phenaki, imagenvideo}, which demonstrate tremendous generalization ability in unseen prompts. 
Although large T2V models yield promising results, they require costly training. To favor affordable video generation, Text2Video-Zero (Zero)~\cite{zero} proposes a zero-shot video generation pipeline without training. Furthermore, to synthesize videos of developing events, works like DirecT2V~\cite{direct2v} and Free-Bloom~\cite{freebloom} leverage Large-Language-Models (LLMs) to synthesize guidance prompts for each frame. 

Despite the marvelous performance of previous zero-shot T2V works, there exist \textbf{two nontrivial challenges}: 
\textit{\textbf{1)} the motion variation pattern remains unaltered and prompt-agnostic, disregarding motion priors inherent in prompts.}
As shown in the video of Zero~\cite{zero} in the top left of Fig.~\ref{frontpage}, because of ignoring motion priors in ``airplane landing" that the airplane moves upwards, the synthesized airplane actually takes off; 
\textbf{2)} the motion control of different objects is inaccurate and entangled without considering independent motion priors of different objects.
For Zero~\cite{zero}, the motion variation strategy is global, which cannot distinguish the dynamic and static parts. 
For LLMs-assisted models~\cite{direct2v, freebloom}, solely relying on prompt guidance for each frame cannot accurately control replacement of objects and incurs incoherence. As depicted in the video of DirecT2V~\cite{direct2v} in Fig.~\ref{frontpage}, frames are fragmented and lack coherence.

To address this issue, we propose a prompt-adaptive and disentangled motion control strategy coined as \textbf{MotionZero} demonstrated in Fig.~\ref{pipeline}, which consists of two major components: Extracting Motion Priors and Disentangled Motion Control.
\textit{In Extracting Motion Priors}, to mimic humans envisioning scenes and corresponding moving trajectories given prompts, we exploit motion information from prompts by querying LLMs which objects will move and in which direction will they move in each video frame. For example, ``A man skateboarding from a skiing mountain" implies a ``down'' moving direction for the ``man'' in each frame. However, for verbs those are not explicitly directional like ``walking'', solely relying on prompts is inadequate. Therefore, we first infer the first frame of the synthesized video using T2I models and feed the first frame to LLMs for accurate moving direction inference.

\textit{In Disentangled Motion Control}, conditioned on the exploited motion information for different objects, we aim to apply the specific moving directions precisely on specific objects. Thus, we adopt an open-domain and prompt-driven segmentation model~\cite{sam} to locate the initial location of moving objects from the first frame.
Then features of subsequent frames in the denoising process will be warped under the guidance of direction and location information.

Besides motion control, existing cross-frame attention schemes, which establish attention among frames for coherence, are also prompt-agnostic. Such attention schemes cannot adapt to different degrees of motion amplitude in different videos. In videos with drastic change like ``skateboarding up'', attending to early frames will restrict the motion change. 
Intuitively, we propose a \textit{Motion-Aware Attention scheme} to determine attending frames according to motion amplitude.

Our MotionZero also empowers versatile video applications as demonstrated in Fig.~\ref{replace} and Fig.~\ref{camera}.
The contributions can be summarized as follows: 

1) We propose a prompt-adaptive and disentangled motion variation strategy for zero-shot Text-to-Video generation, which firstly exploits motion priors of moving objects given prompts by LLMs and then applies the motion priors separately on corresponding objects.

2) To accommodate videos of different motion amplitudes(`run' vs `breeze'), we introduce a Motion-Aware Attention scheme which adjusts attention cross frames according to motion amplitudes.

3) Extensive experiments demonstrate that our method could correctly model and control motion in generated videos and support versatile applications in a zero-shot way.

\section{Related Works}

\subsection{Text-to-Video Synthesis and Edit}
In the early stage of T2V, GAN with 3D-convolutions are utilized to generate videos~\cite{vgft, PanQYLM17, balaji2019conditional}. 
Different from the adversarial process, VDM~\cite{vdm} firstly adopts diffusion models for video generation. 
However, these works are primarily limited in generalization ability. 

Inspired by large-scale T2I models~\cite{dalle2, cogview2, ldm, imagen}, subsequent works also train T2V models on open-domain corpora through transformers~\cite{wu2021godiva, nuwa, Phenaki} or diffusion models~\cite{he2022latent} for generalization.
To fully utilize the knowledge in T2I models, recent methods are established upon pretrained T2I models and introduce temporal modules for robust and efficient T2V training.~\cite{cogvideo, makeavideo, zhou2022magicvideo , VideoFactory, ge2023preserve, xing2023simda, wang2023modelscope, an2023latent}. Specifically, to promote the resolution and coherence of videos, a classic way is to insert super-resolution and interpolation modules in a cascaded way like~\cite{imagenvideo, blattmann2023align, wang2023lavie, zhang2023show}. For further video fidelity enhancement and artifact removal, works like~\cite{li2023videogen, liu2023dual} emphasize dynamics and motion learning. Another line of works focuses on modelling complex dynamics in T2V~\cite{lin2023videodirectorgpt, lian2023llmgroundedvideo, fei2023empowering}, which leverage the knowledge of language models for displacement plans. Given general T2V models, works like~\cite{yang2023probabilistic, jiang2023text2performer, yin2023nuwa} transfer knowledge from general to specific domains.


Despite generalization ability, the above large models are costly in resources.
Therefore, to both enable generalized and resource-friendly T2V, zero-shot T2V is proposed in Text-to-Video Zero (Zero)~\cite{zero} in a training-free way. To further model dynamic scenes, DirecT2V~\cite{direct2v} and Free-Bloom~\cite{freebloom} assign each frame with a prompt drawn from LLMs. Moreover, \cite{duan2023diffsynth} focuses on the issue of flickering. Instead of building on T2I models, LLM-Grounded VDM~\cite{lian2023llmgroundedvideo} leverages both LLM and pretrained video diffusion models to control the spatial variation of objects. 

Aside from the conventional T2V works, a range of applications are studied including video edit~\cite{molad2023dreamix, esser2023structure, tav, liu2023video, qi2023fatezero}, pose-guided video synthesis~\cite{ma2023follow, qin2023dancing}, depth-guided video synthesis~\cite{zhang2023controlvideo}, multimodal-guided video synthesis~\cite{ruan2023mm, wang2023videocomposer, MovieFactory}, animating stories~\cite{he2023animate} and concept customization~\cite{xing2023make}.



\subsection{Leveraging Knowledge from LLMs}
Large Language Models (LLMs) have demonstrated unprecedented abilities on various natural language processing (NLP) tasks like understanding~\cite{brown2020language, ouyang2022training, openai2023gpt4} and chain-of-thought reasoning~\cite{zhou2022least}. 
With recent progress in Embodied AI, commensense knowledge contained in LLMs is extended to planing~\cite{huang2022language, lu2023multimodal} and complex task solving~\cite{shen2023hugginggpt}. 

For visual synthesis works~\cite{brooks2023instructpix2pix, direct2v, freebloom}, LLMs act as directors for giving instructions to text-to-visual models. Also, LLMs act as planners to arrange spatial or temporal layouts for visual synthesis guidance.~\cite{lian2023llm, feng2023layoutgpt, phung2023grounded, lian2023llmgroundedvideo, lin2023videodirectorgpt}

\begin{figure*}[]
\centering
\includegraphics[width=0.95\linewidth]{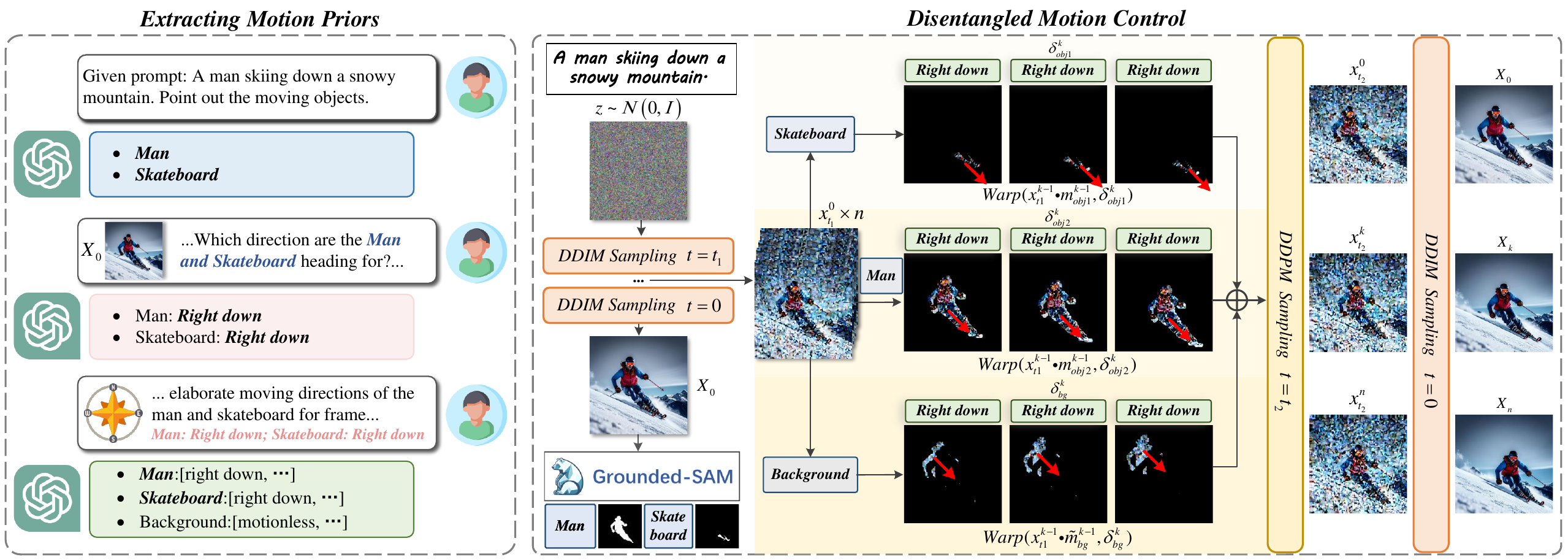}
    \caption{\textbf{Pipeline of our method.} \textit{In Extracting Motion Priors}, motion priors of different objects (moving objects and moving directions) for each frame are inferred by LLMs conditioned on text prompts and the generated first frame. \textit{In Disentangled Motion Control}, the extracted motion priors are respectively applied on corresponding characters in the feature space, located by their masks from SAM~\cite{sam}.}
\label{pipeline}
\end{figure*}

\section{Preliminaries}

\textbf{Denoising Diffusion Probabilistic Models.} 
DDPMs include two Markov chains: a forward chain that adds noises to clean data and a reverse chain that recovers clean data from noises.~\cite{ho2020denoising}
Specifically, in the \textit{forward chain} of $T$ timesteps, noises $\epsilon$ sampled from a Gaussian distribution $\mathcal{N}(0, 1)$ are gradually injected to data $x_0$ to form a Gaussian noise $x_T$ eventually. The forward process $q(x_t|x_{t-1})$ could be written as:
\begin{equation}
q(x_t|x_{t-1}) = \mathcal{N}(x_t; \sqrt{1-\beta_t}x_{t-1}, \beta_tI) , t = 1,...,T
\end{equation}
where  ${\{\beta_i\}}^T_{i=1}$ is designed to enable $q(x_T|x_{0}) \approx \mathcal{N}(0, 1)$. 
While for the \textit{reverse chain}, a time-conditioned U-Net model $\theta$ is trained to predict added noises $\epsilon$ each timestep such that $x_0$ could be converted from $x_T$.
After training, one can sample target data from a Gaussian noise by feeding the noise into the cascaded chained U-Net. Particularly, DDIM~\cite{song2020denoising} serves as a fully deterministic process.

\textbf{Stable Diffusion.} SD~\cite{ldm} is acknowledged as a computationally efficient and powerful pretrained Text-to-Image model, which maps an image $X_0$ in the pixel level into the latents level $x_0$ by using a trained VQGAN encoder $\mathcal{E}$. When projecting back to pixel level, a VQGAN decoder $\mathcal{D}$ decodes $x_0$ back to $X_0$. Then SD utilizes U-Net equipped with attention modules to model a diffusion process. 

\section{Methods}
Given a text prompt $\mathcal{T}$ as the input, our method MotionZero generate cohesive and textual-aligned video frames $\mathcal{V} \in \mathbb{R}^{n \times 3 \times h \times w}$ without any videos for training or finetuning. 
The pipeline of our model MotionZero is illustrated in Fig.~\ref{pipeline}, composed of Extracting Motion Priors in ~\ref{4.1} and Disentangeld Motion Control in Sec.~\ref{4.2}. 

\subsection{Extracting Motion Priors}
\label{4.1}
Given a text prompt, humans could envision the described scene and infer how the scene move. To intimate the behaviors, we exploit motion priors contained in text prompts and the generated first frame for guidance in video generation. Since Large Language Models (LLMs) excel in reasoning, we resort to LLMs for motion priors extraction.

\textbf{Priors in Prompts.} Specifically, we assume that motion information in videos could be dissected into different parts moving towards different directions in each frame. For example, given the prompt ``a man skiing down the snowy mountain", the ``man'' will keep skiing ``downward'' in each frame.
To obtain the motion information, we enable LLM firstly to identify the moving parts and motionless parts given text prompts, as shown in the left of Fig.~\ref{pipeline}. 
Given the moving parts, we could further design prompts to extract their moving directions. To facilitate the representation of directional information, we partition the coordinate axes into eight directional motions \textit{[`up', `down', `left', `right', `left up', `left down', `right up', `right down']}. Conditioned on the direction coordinates, text prompts and moving parts, LLM could elaborate the moving directions of different moving parts of each frame. 

\textbf{Priors in First Frames.} However, for motion that is not explicitly directional, solely relying on information in prompts is not adequate. For example, `walking' or `riding bikes' are not specifically directional compared to `jumping' or `landing'. 
Beside prompts, humans usually envision event development based on the initial scene. Inspired by this habit, we utilize vision information contained in the first frame of the synthesized video to infer more accurate moving directions. In details, we firstly generate the first frame $X_0$ given text prompt $\mathcal{T}$ and random noise sampled from Gaussian distribution $z_T \sim \mathcal{N}(0, 1)$ by deterministic DDIM sampling.
\begin{equation}
    X_0 = \mathcal{D}(DDIM_{t=0}(z_T, \mathcal{T})).
\end{equation}
With $X_0$, we select Visual Question Answering models to infer the heading directions of the moving parts. 

With the heading directions as priors from $X_0$, the predefined coordinates and input prompts, LLMs are asked to reason moving directions of each moving part for each frame. For the example of skiing man, the final motion information will be illustrated as \textit{\{Man:[right down, ...], ...\}}. While the other parts are taken as motionless background.


\subsection{Disentangled Motion Control}
\label{4.2}
\begin{figure*}[]
\centering
\includegraphics[width=1\linewidth]{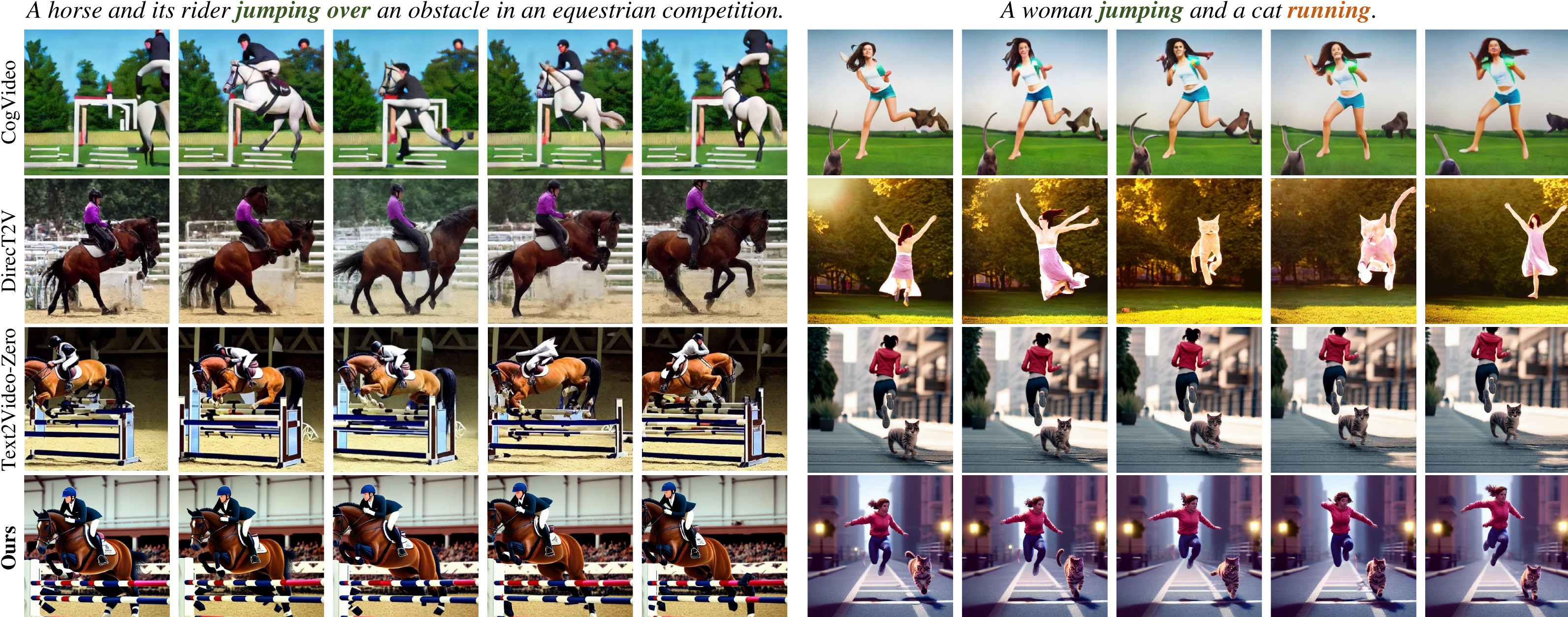}
\caption{\textbf{Visual results comparison with CogVideo~\cite{cogvideo}, DirecT2V~\cite{direct2v} and Text2Video-Zero~\cite{zero}.} \textit{Left: }Ours achieves accurate motion control with background preservation and video coherence. \textit{Right: }For multiple objects movement, we could separately control different objects with corresponding motion.}
\label{comparison_1}
\end{figure*}

Given extracted motion priors, we propose to directly apply them on moving objects, avoiding inaccurate control by prompt guidance.
Though ~\cite{zero} also directly applies motion information, it globally warps the whole features of frames. 
Consequently, there is no relative movement as shown in the third row of Fig.~\ref{comparison_1}, which violates motion regulation.
Therefore, background and moving characters should be disentangled controlled. Moreover, different characters also require disentangled control, since they follow different moving patterns.

\textbf{Spatial Location of Objects.} To achieve disentangled control, we are supposed to firstly locate spatial regions of different characters.
As observed~\cite{tumanyan2023plug} in diffusion models, regions in the pixel level spatially correspond to the regions in the feature level. Based on the observation, we utilize mature segmentation tools~\cite{sam, liu2023grounding} driven by prompts to acquire initial location of objects. As shown in Fig.~\ref{pipeline}, given the first frame $X_0$ and the prompts $\mathcal{T}_i$ of $i$-th moving character like ``man'', Grounded-SAM $SAM(\cdot)$ outputs corresponding masks as:
\begin{equation}
    m^{0}_{i} = SAM(X_0, \mathcal{T}_i) \indent i=(1, ..., c),
\end{equation}
where $c$ indicates the total amount of the moving characters and $m^0_i$ refers to the initial location of the $i$-th character in the first frame.

\textbf{Motion Control of Objects.} Based on the motion priors of different characters like \textit{\{Man:[right down, ...]\}} and their corresponding location, we could conduct motion variation precisely on different moving characters.
As shown in Fig.~\ref{pipeline}, motion priors are applied on latent features of different frames, which are initialized to be the feature of the first frame. Specifically, in the $t_1$ timestep of the DDIM sampling process, the feature of the first frame $x_{t_1}^0$ is formulated as:
\begin{equation}
    \begin{aligned}
    x_{t_1}^0 &= DDIM_{t=t_1}(z_0, \mathcal{T}) \\
    x_{t_1}^{k} &= x_{t_1}^0 \indent k=(1,...,n).
    \end{aligned}
\end{equation}

To ensure the coherence among frames, motion of subsequent frames should be modified based on preceding frames. Thus, motion variation of the man skiing right down in the $k$-th frame is represented as moving the man in the $(k-1)$-th frame towards right down more. Specifically, the features indicated by the mask of the man $m^{k-1}_i$ in the $(k-1)$-th frame will be warped towards right down. Similar warp operations are also conducted on features of other moving characters, which are finally summed over as the feature of the $k$-th frame. 
The process is formulated as:

\begin{equation}
    \begin{aligned}
    x_{t_1}^{k} = \sum_{i=1}^{c+1}& [ Warp(x_{t_1}^{k-1}, \delta_i^k)  \\ \cdot &Warp(m_i^{k-1} \cup \widetilde{m}_i^{k-1}, \delta_i^k) ] \\ 
    \widetilde{m}_i^{k-1} = &Warp(m_i^{k-1}, -\delta_i^k) \\
    m_i^{k} = &Warp(m_i^{k-1}, \delta_i^k),
    \end{aligned}
\label{warp}
\end{equation}
where $\delta_i^k$ refers to motion directions of the $i$-th character in the $k$-th frame. $m_i^{k-1}$ is the mask indicating the region of the $i$-th character in the $k$-th frame. And $\widetilde{m}_i^{k-1}$ represents $m_i^{k-1}$ moving towards the reverse direction to include the background in the reverse direction. Therefore, $m_i^{k-1} \cup \widetilde{m}_i^{k-1}$ altogether moving could hide the original trace indicated by $m_i^0$ like image inpainting with surrounding background. 
Notably, the $(c+1)$-th character refers to the motionless background except for all other moving characters. So the feature of the background will not be warped and is set identical to the first frame.

To avoid rigid ``cut and paste'' visual effect, the modified feature $x_{t_1}^{k}$ will go through DDPM sampling process to introduce stochastic fusion and variation. After DDPM sampling of several timesteps, the modified features are then fed into DDIM sampling to obtain the final video features $x_{0}^{1:n}$. And the final videos $\mathcal{V}$ could be obtained through decoding of the video features. We formulate the process as follows:


\begin{equation}
    \begin{aligned}
    x_{t_2}^{1:n} &= DDPM_{t=t_2}(x_{t_1}^{1:n}, \mathcal{T}) \\
    x_{0}^{1:n} &= DDIM_{t=0}(x_{t_2}^{1:n}, \mathcal{T}) \\ 
    \mathcal{V} &= \mathcal{D}(x_{0}^{1:n})
    \end{aligned}
\end{equation}


\subsection{Motion-Aware Attention}
\label{4.3}
Cross Frame attention are vital in ensuring dynamics and frame coherence. In previous zero-shot T2V works, Cross Frame attention either keeps attending to the first frame~\cite{zero} or variates along with timesteps~\cite{direct2v, freebloom}. All of the above attention schemes are text-irrelevant. 

While different prompts describe motion of different dynamic amplitude. For example, a prompt like ``a girl skateboarding high'' implies drastically changed motion or displacement. While a prompt like ``man writing'' or ``the brook murmurs'' depicts a relatively less dynamic scene. Particularly, largely dynamic scenes require large motion variation, still attending to the first frame will hallucinate the displacement and pose of characters, as shown in the fourth row of Fig.~\ref{ablation}. And for quiet scenes, motion change is relatively limited, attending to early frames will contribute to semantics preservation. 

Inspired by this, we propose a Motion-Aware Attention scheme, which is adaptive according to the motion amplitude. 
To judge the scale of motion amplitude, we employ \textit{Intersection over Union} (IoU) as metrics to determine whether the movement between frames is overly large. Specifically, we define a frame that subsequent frames attend to as the \textit{Anchor Frame}. Initially, the first frame is taken as the Anchor Frame. Naturally, IoU is calculated between the mask of the current frame $m^{k}$ and the mask of the Anchor Frame $m^{a}$. 
When the IoU is below a specific threshold, it implies that the current frame is largely deviated from the Anchor Frame and the Anchor Frame needs update.
Otherwise, the current frame is still visually similar to the Anchor Frame and attending to the Anchor Frame could assure coherence. 


As described above, the Motion-Aware Attention $MAA(\cdot)$ is formulated as follows:

\begin{equation}
   MAA(Q^k, K^{1:n}, V^{1:n}) = Softmax(\frac{Q^k(K^{a})^T}{\sqrt{d}})V^{a} 
   \label{attention}
\end{equation}

\begin{equation}
a =
\begin{cases}
\centering
k-1 &\quad 1 - IoU(m^{a}, m^{k}) \le \gamma\\
a &\quad 1 - IoU(m^{a}, m^{k}) > \gamma,
\end{cases}
\end{equation}
where $\gamma$ is set as the threshold for IoU evaluation, which controls the sensibility to motion amplitude. And the initial $a$ is set as $0$ to preserve visual content of the first frame.

\section{Applications}

\begin{figure}
    \centering
    \includegraphics[width=0.9\linewidth]{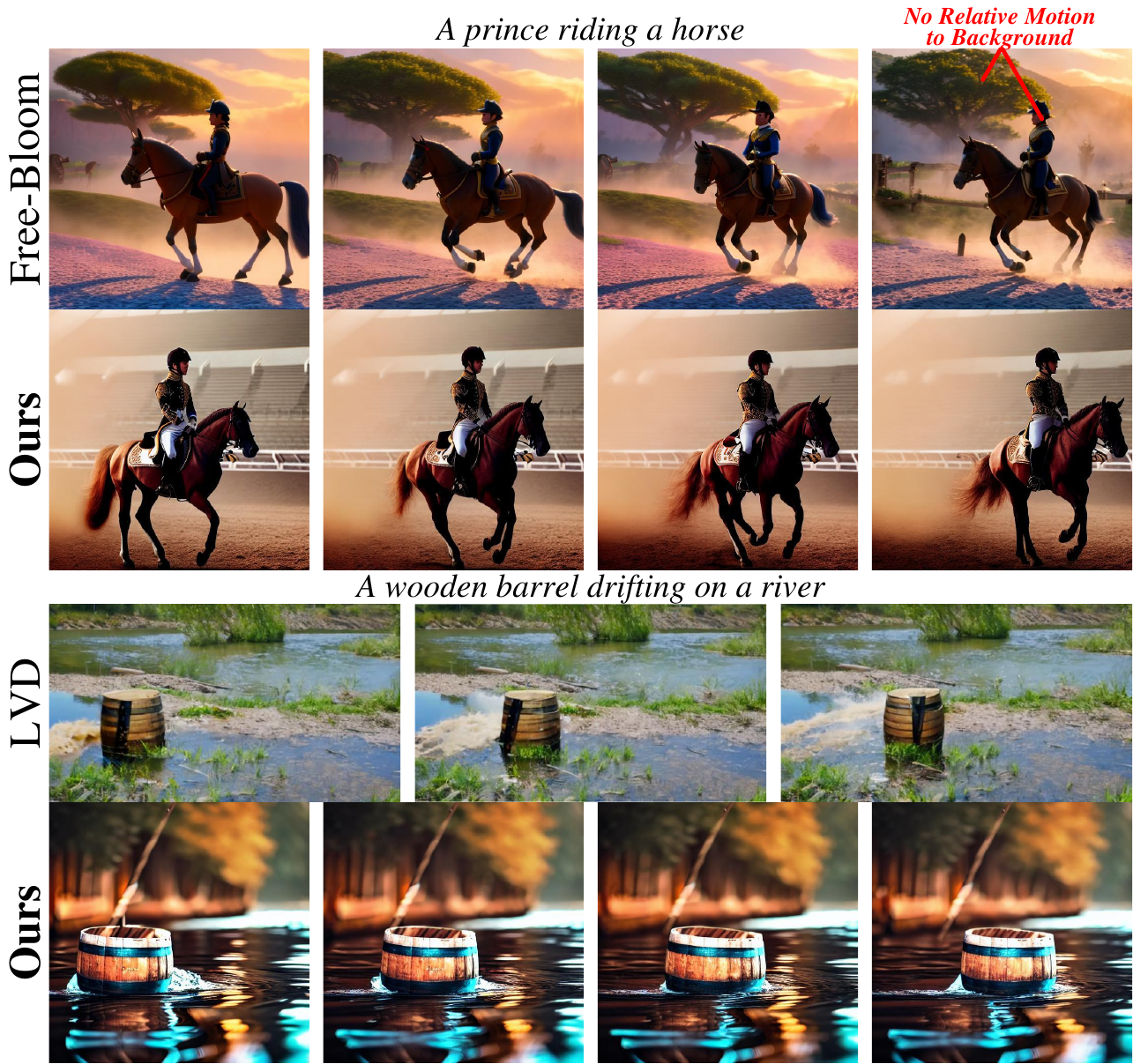}
    \caption{\textbf{Visual results comparison with concurrent works Free-Bloom~\cite{freebloom} and LVD~\cite{lian2023llmgroundedvideo}.} Ours performs better than Free-Bloom in displacement and is comparable with large models LVD.}
    \label{fig:free_bloom}
\end{figure}

\begin{figure}[]
\centering
\includegraphics[width=1\linewidth]{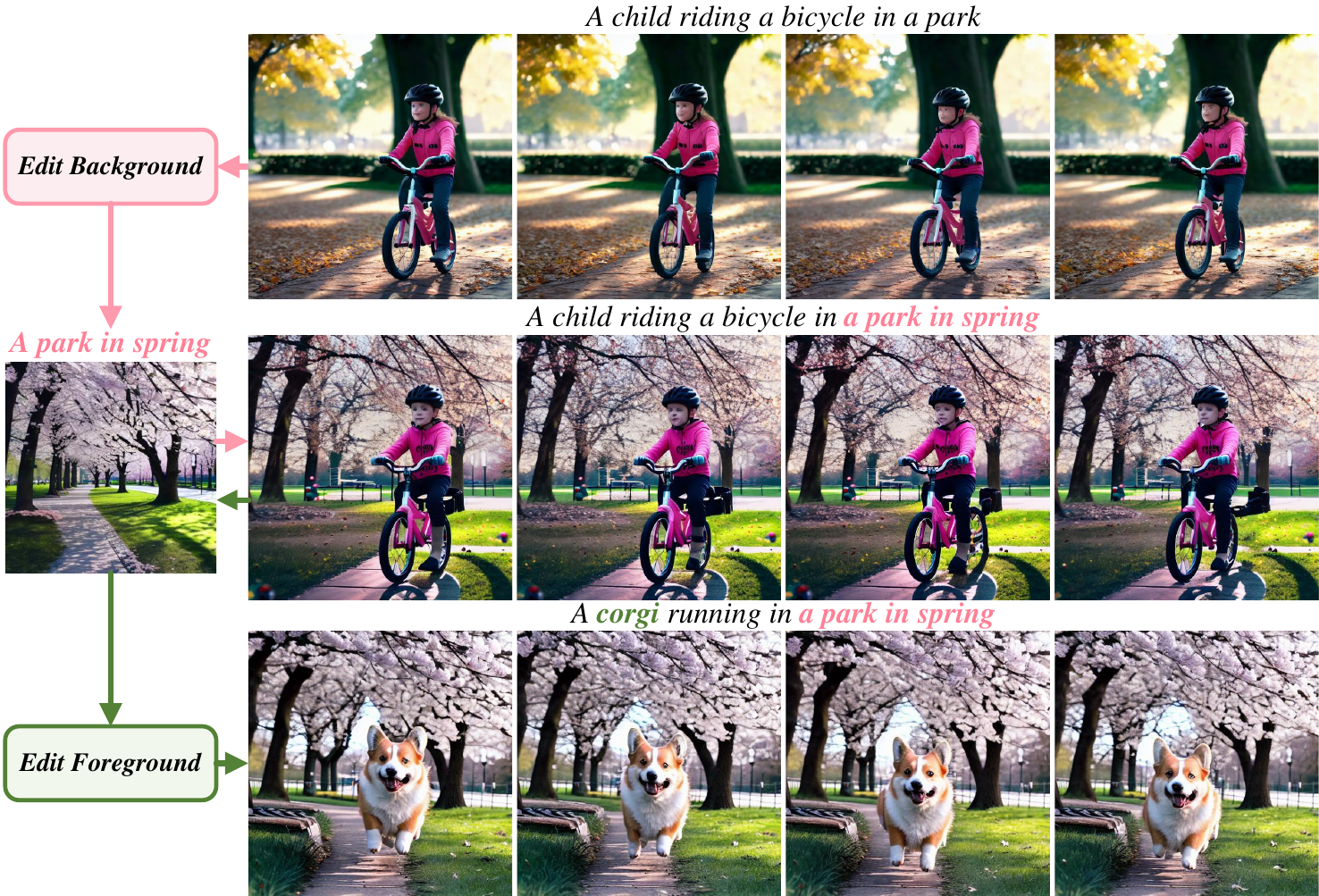}
\caption{\textbf{The application of editing foreground and background.} Given a synthesized video, our Disentangled Motion Control allows for separately editing foreground and background semantically.}
\label{replace}
\end{figure}

\textbf{Edit Background and Foreground} (Fig.\ref{replace}) Since different parts of videos are independently controlled, our method naturally allows for semantically editing foreground and background in a zero-shot way. 
Given background features $x^{bg}_{t_{1}}$ and foreground features $x^{fg}_{t_{1}}$ respectively controlled by background prompt $\mathcal{T}_{bg}$ and foreground prompt $\mathcal{T}_{fg}$, we utilize the mask $m^{fg}$ of the foreground to fuse the two features as follows:
\begin{equation}
    x_{t_1} = x^{fg}_{t_{1}} \cdot m^{fg} + x^{bg}_{t_{1}} \cdot (1-m^{fg}).
\end{equation}
As shown in the third row of Fig.~\ref{replace} for replacing the foreground, the semantics of the background are largely preserved and naturally interacted with the foreground. 

\begin{figure*}[]
\centering
\includegraphics[width=0.95\linewidth]{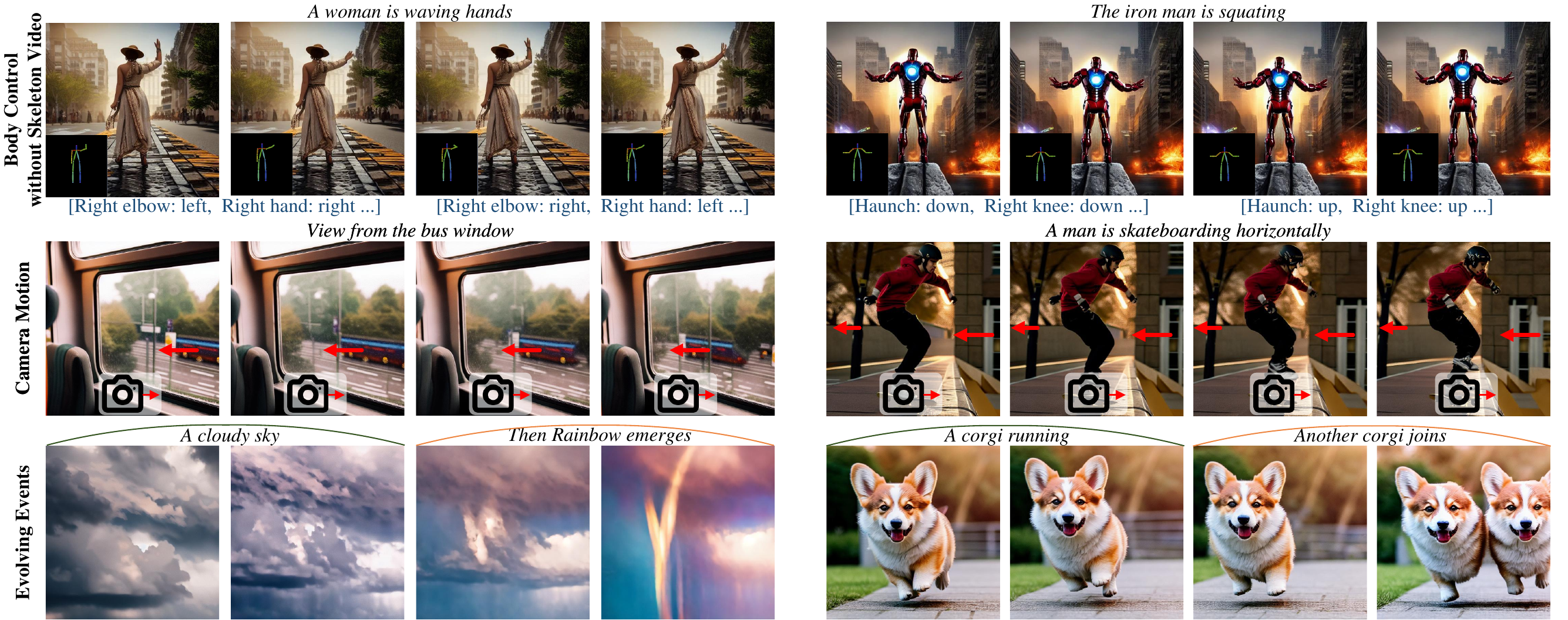}
\caption{\textbf{Different applications based on our MotionZero.} \textit{1) Body Control without Skeleton Video}(1-2 row) Skeleton movement information is extracted from LLMs to synthesize a skeleton video, which then guides video generation; \textit{2) Camera motion}(3-4 row) Camera following the protagonist implies a movement in the reverse direction(backward scene); \textit{3) Evolving Events}(5-6 row) Contextually related prompts control different slices of a video.} 
\label{camera}
\end{figure*}

\textbf{Body Control without Skeleton Video} (Row 1 of Fig.~\ref{camera})
To facilitate more sophisticated control over human body, our Extracting Motion Priors could also support exploiting motion information about skeleton. Different from ~\cite{zero, ma2023follow} requiring skeleton videos, only text prompts are needed. Particularly, we provide LLMs with priors that the skeleton represents human joints or parts with 10 discrete nodes. Then, given prompts like ``waving hands'', LLMs could infer the movement of each node for each video frame like \textit{``Frame 2: right hand: right''}. Consequently, a skeleton video of ``waving hand'' could be generated. With the aid of ControlNet~\cite{zhang2023adding} and our video generation pipeline, corresponding videos could be generated as demonstrated in Fig.~\ref{camera}. 

\textbf{Camera Motion} (Row 2 of Fig.~\ref{camera})
Camera following the protagonist is also a common filming technique. Considering the relative motion, the background looks like moving towards the reverse direction. Therefore, motion information of the foreground will be inversely applied to the background through Disentangled Motion Control.

\textbf{Evolving Events} (Row 3 of Fig.~\ref{camera})
When given prompts imply a multi-stage development or contain multiple sentences, videos could be naturally divided into slices according stages or sentences.  For example, ``cloudy day turns to rainbow" indicates two-stages which are ``cloudy day'' first and ``rainbow emerges'' secondly. In such way, different prompts control different slices, where the Anchor Frames are independently controlled in each slice.

\section{Experiments And Analysis}
\subsection{Evaluation Metrics}

\textbf{1) Textual Alignment.} We utilize CLIP~\cite{clip} to calculate the cosine similarity between visual features of frames and textual features of prompts.
\textbf{2) Motion Correctness. } 
CLIP can only evaluate semantics alignment of single frames with prompts, ignoring motion variation among frames.
Therefore, we adopt video tracking models~\cite{cheng2023segment} to obtain the moving trajectories of characters in synthesized videos, which could reflects motion variation. Then we calculate the accuracy between the obtained trajectories and given trajectories drawn from LLMs. To avoid unfair comparison, we select 10 prompts with obvious heading direction information like ``landing".
\textbf{3) User Preference. }For human perception reflection, we invite 20 human raters from universities to mark their preferences between two baselines. We ask users to judge preferences from the perspectives of text-visual alignment and overall video quality.

\subsection{Comparisons}

\begin{table}[]
\resizebox{\linewidth}{!}{\begin{tabular}{lcclcc}
\hline
\multirow{2}{*}{Methods} & \multicolumn{2}{c}{\textit{Qualitative Results}}                                                                           & \multirow{2}{*}{Methods} & \multicolumn{2}{c}{\textit{User Preference}}                                                                          \\ \cline{2-3} \cline{5-6} 
                         & \begin{tabular}[c]{@{}c@{}}Textual\\ Alignment\end{tabular} & \begin{tabular}[c]{@{}c@{}}Motion\\ Correctness\end{tabular} &                          & \begin{tabular}[c]{@{}c@{}}Textual\\ Alignment\end{tabular} & \begin{tabular}[c]{@{}c@{}}Video\\ Quality\end{tabular} \\ \hline
CogVideo                 & 28.10                                                       & 55.71\%                                                      & vs CogVideo         & 80.5\%                                                      & 81.0\%                                                  \\
DirecT2V                 & 27.93                                                       & 20.00\%                                                      & vs DireT2V          & 95.0\%                                                      & 89.0\%                                                  \\
Zero                     & 29.95                                                       & 35.71\%                                                      & vs Zero             & 79.5\%                                                      & 81.0\%                                                  \\
Ours                     & \textbf{30.25}                                              & \textbf{82.86\%}                                             & /    & /                                                           & /                                                       \\ \hline
\end{tabular}}
\caption{\textbf{Qualitative comparisons among CogVideo, DirecT2V, Text2Video-Zero and Ours.} In Model Evaluation, CLIP score measures Textual Alignment, and Motion Correctness detects whether characters move as prompt orders. In User Preference, `vs CogVideo' denotes the preference of ours over CogVideo.}
\label{comparison}
\end{table}

\begin{figure*}
    \centering
    \includegraphics[width=1\linewidth]{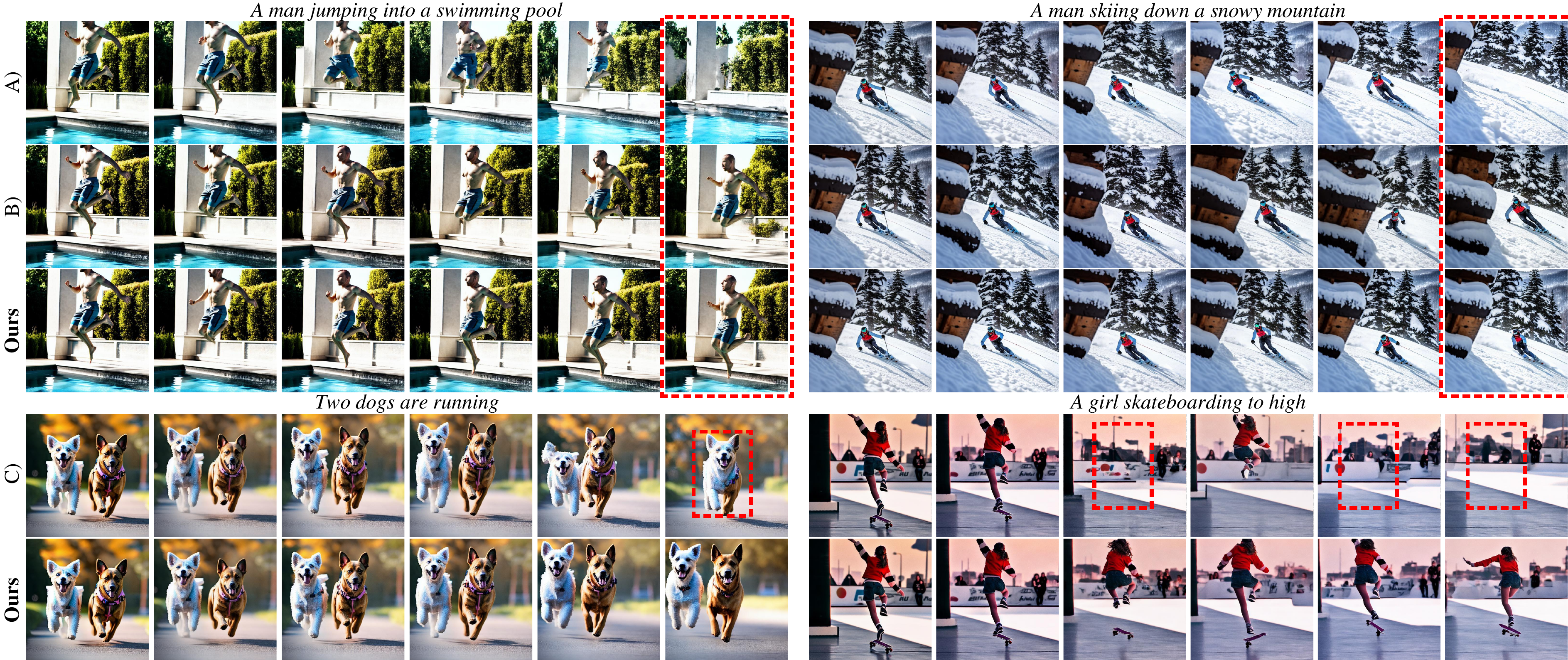}
    \caption{\textbf{Ablation study on the effect of our proposed components.} A) Without Extracting Motion Priors. B) Without Disentangled Motion Control. C) Without Motion-Aware Attention.}
    \label{fig_ablation}
\end{figure*}

\textbf{Qualitative Comparisons}
To direct reflect visual results, we conduct qualitative comparisons among baselines including CogVideo~\cite{cogvideo}, DirecT2V~\cite{direct2v}, Text2Video-Zero~\cite{zero} as demonstrated in Fig.~\ref{comparison}. We choose two representative prompts for visualization. \textbf{First}, to demonstrate motion correctness, we choose prompt ``A horse and its rider jumping over an obstacle in an equestrian competition''. 
Frames of CogVideo performs poor in semantics, though being trained on large-scale videos. Frames of DirecT2V are drawn from its paper, where the horse does not jump over the obstacle. The results reflect that solely relying on prompts cannot accurately control motion. 
Frames of Text2Video-Zero demonstrate a global movement and distortion in horse heading. 
In contrast, our results achieve the right motion ``jumping over obstacle" with background semantically consistent, which proves the effectiveness of ours in accurate motion control.
\textbf{Second}, ``A woman jumping and a cat running" models multiple characters movement. Although frames of CogVideo demonstrate vivid dynamics, the visual quality is unsatisfying. Especially, frames of DirecT2V are inconsistent since multiple objects generation through T2I models are unstable. For Text2Video-Zero, the motion of the cat and woman is entangled. While ours disentangled control precisely applies the motion of jumping and running on corresponding regions.

\begin{figure}
    \centering
    \includegraphics[width=1\linewidth]{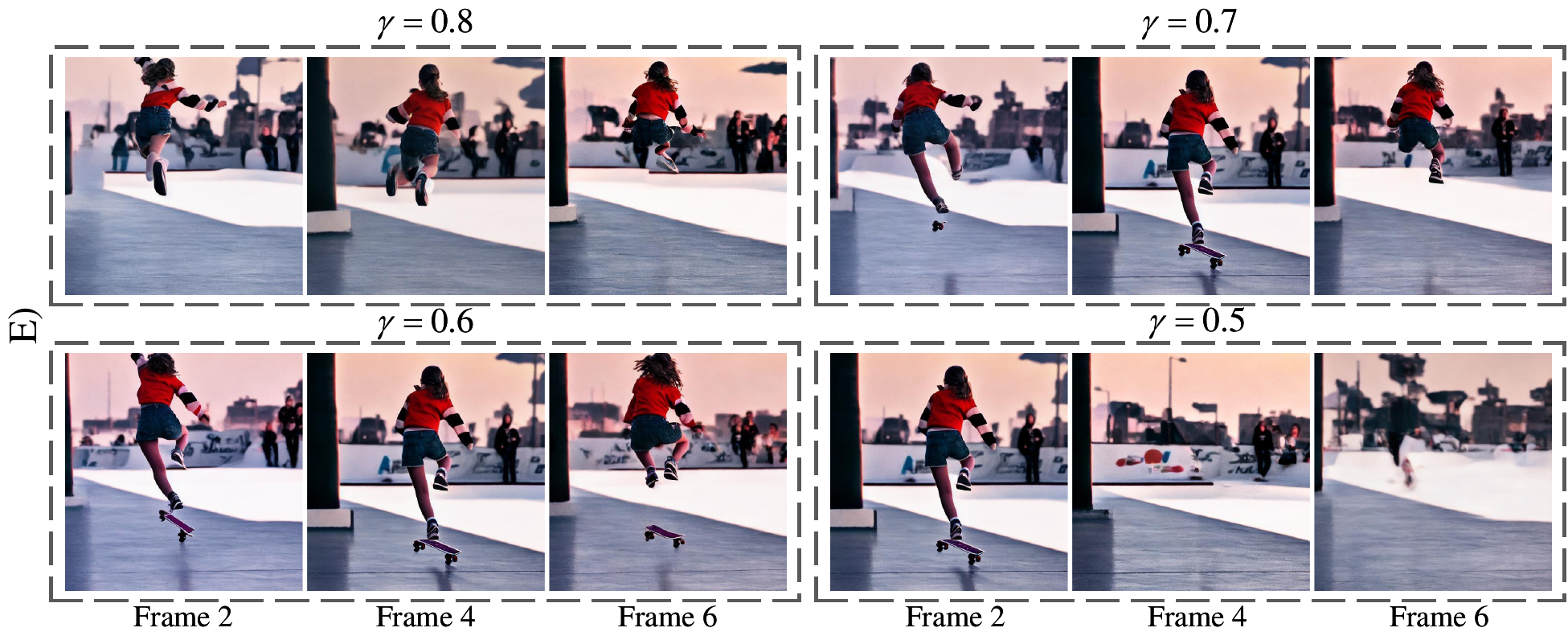}
    \caption{\textbf{Effect of different choices of $\gamma$.}}
    \label{fig:gamma}
\end{figure}

As shown in Fig.~\ref{fig:free_bloom}, we also present visual results compared with concurrent works Free-Bloom~\cite{freebloom} and LVD~\cite{lian2023llmgroundedvideo}, which are obtained from original papers for the absence of published code. Compared with Free-Bloom whose motion priors are implicitly embedded in prompts, ours performs better in relative movement for directly applying motion priors on features. Moreover, ours are comparable with LVD which utilizes pretrained video models.

\textbf{Quantitative Comparisons}
For objective evaluation, we conduct a quantitative comparison among baselines including CogVideo~\cite{cogvideo}, DirecT2V~\cite{direct2v}, Text2Video-Zero~\cite{zero} and Ours as shown in \textit{Model Evaluation} of Tab.~\ref{comparison}. And 25 prompts are randomly selected as inputs for T2V. 

\textit{Textual Alignment.} Video frames are fed into CLIP for text-visual similarity evaluation. 
Particularly, our model achieves the best even compared with a large-scale pretrained T2V model CogVideo. 
DirecT2V performs the worst in text-visual alignment because of the limitation of T2I models in understanding complex prompts. This also prove that imbuing motion information in prompts is inaccurate and uncontrollable.

\textit{Motion Correctness.} However, Clip score could only reflect semantics while fail to reflect motion variation. Thus, we detect motion trajectories among frames and compare them with the given motion trajectories to calculate accuracy depicted as \textit{Motion Correctness} in the Tab.~\ref{comparison}. As shown, our method respectively surpasses CogVideo, DirecT2V and Text2Video-Zero with dominant gap by $27.15$, $62.86$, and $47.15$, which verifies the superiority or ours in motion control. Since being trained on real videos, CogVideo excels at modelling motion information and achieves the second best. DirecT2V gains the worst scores since semantics gaps between texts cause incoherence in adjacent frames.
For Text2Video-Zero, only when given motion trajectories coincides with its fixed motion trajectories will it get right prediction.

\textit{User Preference.} To reflect human real perception, we invite users to rate their preferences between baselines with ours from the perspective of Textual Alignment and Video Quality as shown in the User Preference of Tab.~\ref{comparison}. Specifically, Textual Alignment refers to the alignment between input prompts and the whole video and Video Quality indicates an overall evaluation of the video. Obviously, ours wins others especially DirecT2V by a large margin, which proves our results satisfy users comprehensively.

\subsection{Ablation Study}

\textit{A) Effect of Extracting Motion Priors.}
As shown in the A) of Fig.~\ref{fig_ablation}, we remove Extracting Motion Priors, leading to text-agnostic motion variation(moving upwards for ``jumping into").
\textit{B) Effect of Disentangled
Motion Control.} The B) of Fig.~\ref{fig_ablation} demonstrates the effect of removing Disentangled Motion Control, which directly applies motion priors information on all the pixels. As seen, the background bushes and swimming pool also move downwards without relative motion with the man. \textit{C) Effect of Motion-Aware Attention.}
As demonstrated in C) of Fig.~\ref{fig_ablation}, we abandon our attention scheme and attending all frames to the first frame. By adjusting attention according to motion intensity, our method avoids object distortion like the two merging dogs and object missing like the missing girl.

\textit{E) Effect of different $\gamma$.} As demonstrated in Fig.~\ref{fig:gamma}, we conduct four experiments by respectively setting $\gamma$ as 0.8, 0.7, 0.6 and 0.5 to reflect the visual results along with IoU thresholds.
When $\gamma$ is relatively large as 0.8, the appearance consistency is worsened because of frequently changing the anchor frame. When $\gamma$ is relatively small as 0.5, the update of the anchor frame does not follow the motion change of objects, resulting in the propagation of missing object. When $\gamma=0.5$, approximately $\frac{1}{3}$ area of one object is deviated from the other object. Therefore, updating the anchor frame when IoU is below 0.5 is too late to follow motion variation.
From visual results, we can tell that when $\gamma = 0.6$ videos keep both visually consistent and coherence.

\section{Conclusion}
In this paper, we propose a prompt-adaptive and disentangled motion control strategy MotionZero for zero-shot Text-to-Video synthesis, which exploits motion priors for different objects from prompts and generated first frames. Given the priors, disentangled motion control are applied on corresponding objects. 
Extensive experiments and applications like zero-shot video edit prove the controllability and versatility.

{\small
\bibliographystyle{ieee_fullname}
\bibliography{egbib}
}

\newpage
\begin{appendices}

We provide implementation details to enhance the reproducibility of our approach in Sec.~\ref{details}. And additional visual results are also provided to further support our contributions in Sec.~\ref{visual}. Moreover, additional ablation studies are presented in Sec.~\ref{ablation_add}.
We will release our core code at \url{https://anonymous.4open.science/r/MotionZero-4D75}.


\section{Details}
\label{details}
\subsection{Implementation Details}

Our zero-shot video synthesis model is extended upon text-to-image synthesis model Stable Diffusion~\cite{ldm}. For large language model, we select GPT4~\cite{openai2023gpt4} for the sake of effectiveness and feasibility. And we utilize open-domain prompt-driven for segmenting masks of moving characters~\cite{kirillov2023segany, liu2023grounding}. The confidence of segmenting is set as 0.3. All image sizes are set as $512 \times 512$, and $8$ frames for each video. Additionally, all experiments are conducted on NVIDIA Tesla V100 GPU. Inferring each video takes about 2 minutes. 

\subsection{Details of Interaction with LLMs}
We harness the power of LLMs to reason the motion priors contained in prompts by the following commands:

\textbf{Moving Objects: }\textit{Given a user prompt, identify the moving objects or parts. Prompt: "An airplane is landing on the runway."}

\textbf{Moving Directions: }\textit{Given a user prompt, elaborate the movement direction of the main characters or moving parts for each two frames of 8 frames. Motions should be consistent. Directions should be one of followings:[``motionless", ``left”, ``right”, ``up”, ``down”, ``left down", ``left up", ``right down", ``right up"]. Answer the question in a list follow the format:[``character name": ``direction", ``part name": ``direction", ...].
Prompt: ``An airplane is landing on the runway." And the airplane is heading towards left.}
The priors of ``heading towards left" are drawn from Visual Question Answering models with the input of the first frame.

\subsection{Details of Motion Correctness}
\begin{figure}
    \centering
    \includegraphics[width=1\linewidth]{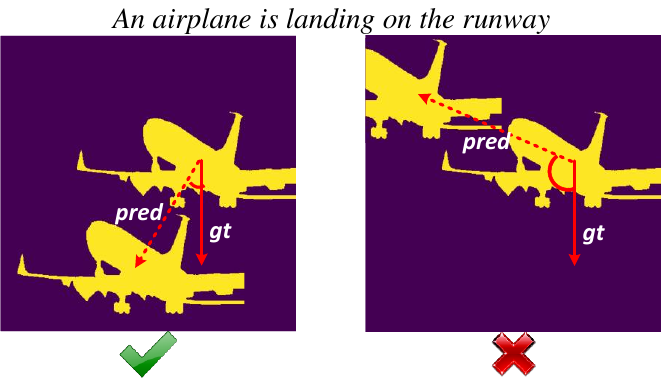}
    \caption{\textbf{Illustration of Motion Correctness.} When cosine similarity between  predicted moving trajectories and ground truth moving trajectories is positive, we judge its motion as correct.}
    \label{fig:motion_correct}
\end{figure}

Given ground truth moving trajectories of specific moving objects for each frame, we compare them with moving trajectories of the synthesized videos to judge the motion correctness as illustrated in Fig.~\ref{fig:motion_correct}.

Specifically, the moving trajectories could be decomposed into each frame moving a certain distance in a specific direction. Here, we only calculate the correctness of moving directions of each frame.
For the $i$-th frame, the ground truth moving direction is $\vec{e}_{gt}^{i}$.
To obtain the moving direction of synthesized videos, we employ an open-domain segmentation model~\cite{kirillov2023segany, liu2023grounding} to track the specific moving objects in each frame. After obtaining the segmentation masks for the moving objects, we set the center point of the bounding box corresponding to each mask as the center point of the respective moving object.
Subsequently, the vector between the center point of the $(i+1)$-th frame and that of the $i$-th frame is regarded as the predicted moving direction $\vec{e}_{pred}^{i}$. 
Ultimately, if the cosine similarity between $\vec{e}_{gt}^{i}$ and $\vec{e}_{pred}^{i}$ is positive, we consider the motion of $i$-th frame as correct. 
\begin{equation}
\begin{cases}
\centering
correct &\quad cos(\vec{e}_{gt}^{i}, \vec{e}_{pred}^{i}) > 0\\
wrong &\quad cos(\vec{e}_{gt}^{i}, \vec{e}_{pred}^{i}) \leq 0,
\end{cases}
\end{equation}
Finally, the average accuracy of different videos are regarded as the final evaluation metric.

\section{Additional Visual Results}
\label{visual}
\subsection{Comparison Visual Results}
\begin{figure*}
    \centering
    \includegraphics[width=0.9\linewidth]{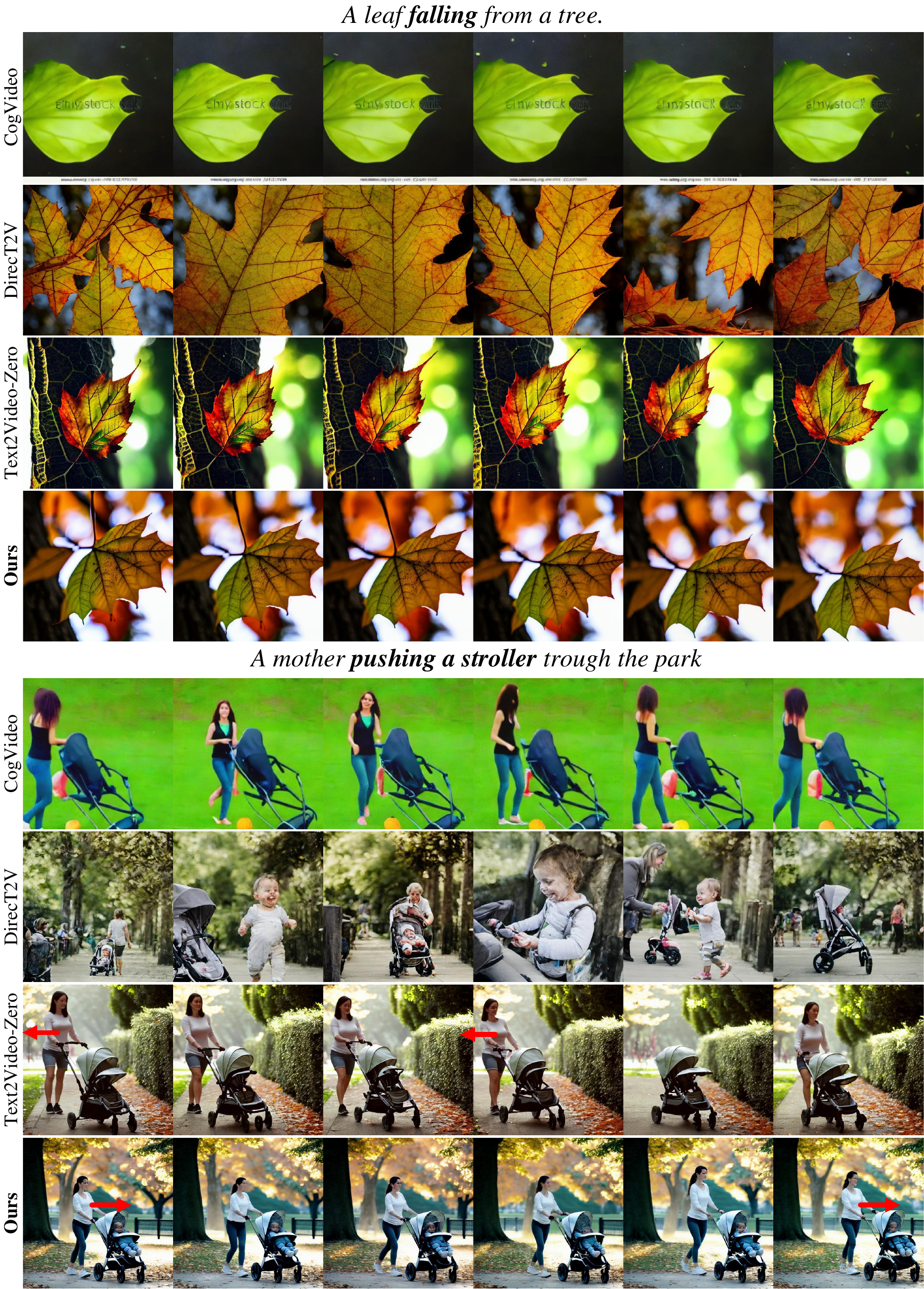}
    \caption{\textbf{Additional comparison visual results with CogVideo~\cite{cogvideo}, DirecT2V~\cite{direct2v} and Text2Video-Zero~\cite{zero}.}}
    \label{fig:comp1}
\end{figure*}

\begin{figure*}
    \centering
    \includegraphics[width=0.9\linewidth]{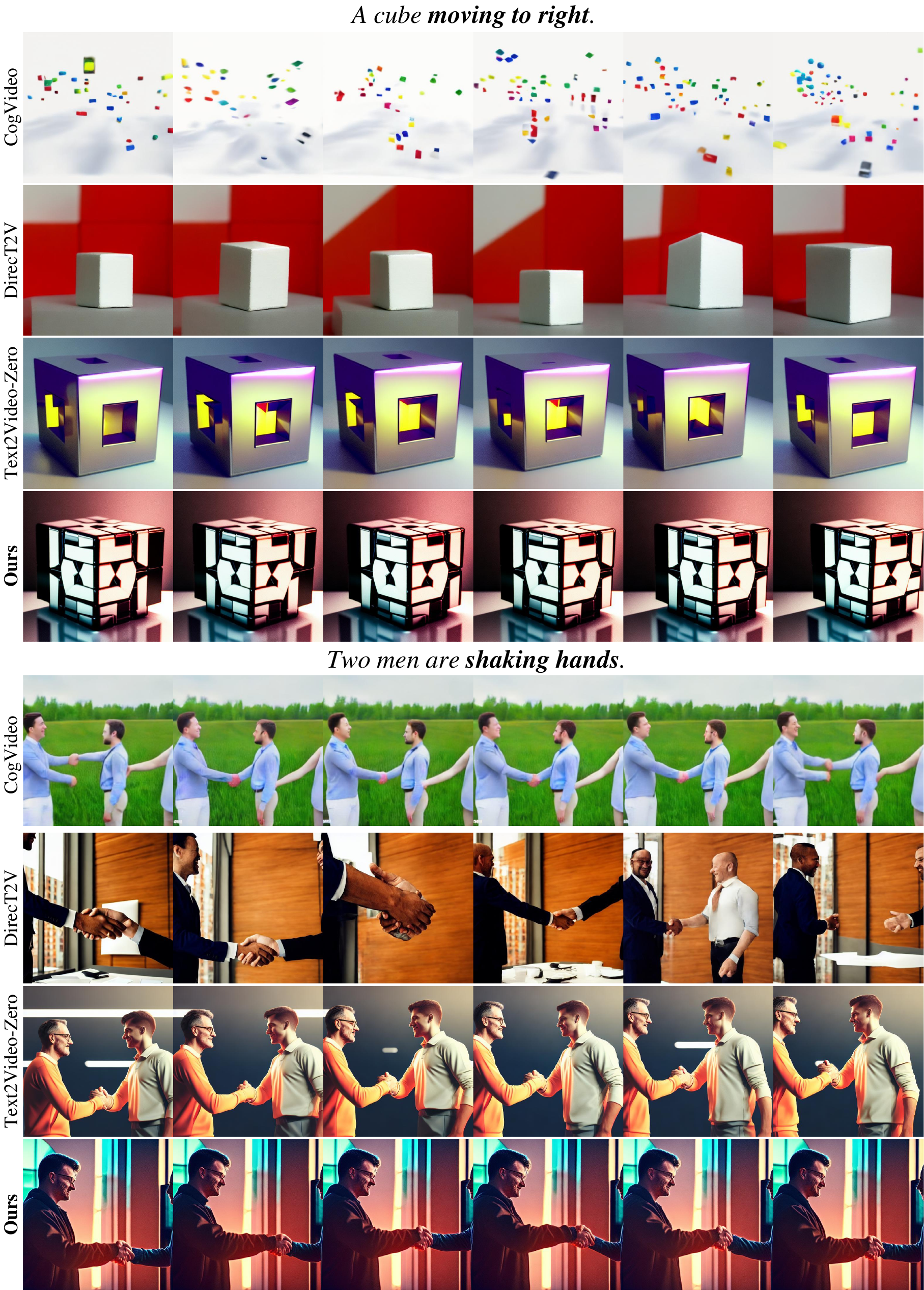}
    \caption{\textbf{Additional comparison visual results with CogVideo~\cite{cogvideo}, DirecT2V~\cite{direct2v} and Text2Video-Zero~\cite{zero}.}}
    \label{fig:comp2}
\end{figure*}

As shown in Fig.~\ref{fig:comp1} and Fig.~\ref{fig:comp2}, we provide additional comparison visual results with CogVideo~\cite{cogvideo}, DirecT2V~\cite{direct2v} and Text2Video-Zero~\cite{zero}. In terms of semantic alignment, motion correctness and consistency, ours achieve the best.

\begin{figure*}
    \centering
    \includegraphics[width=1\linewidth]{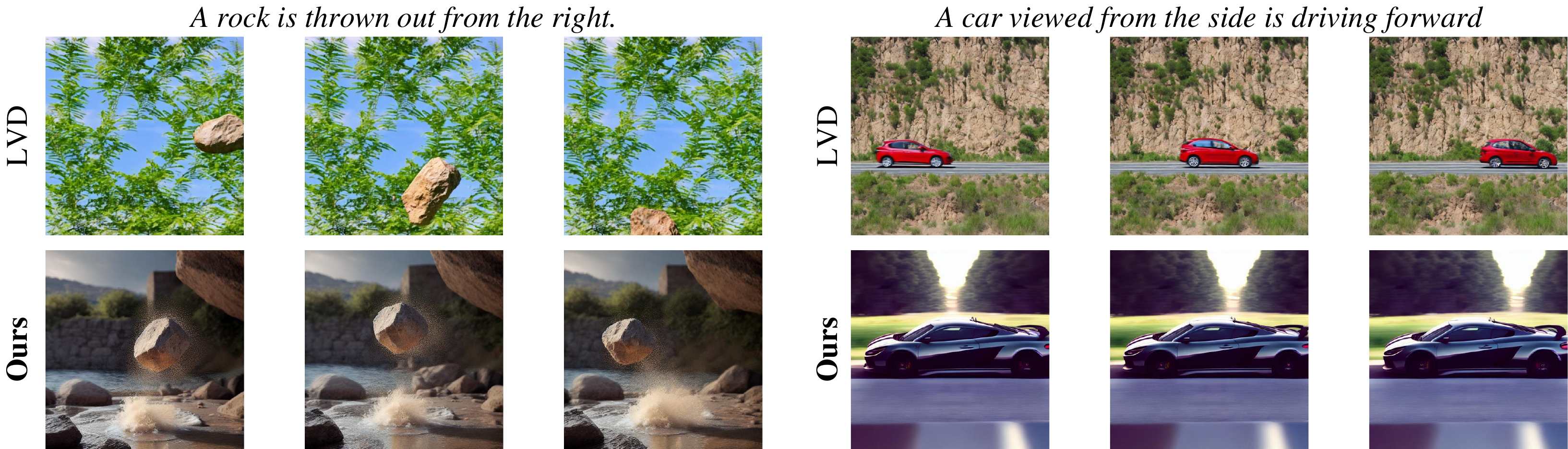}
    \caption{\textbf{Additional comparison visual results with LVD~\cite{lian2023llmgroundedvideo}.} Except for concise replacement control, ours perform better in object appearance consistency, e.g.,the ``rock".}
    \label{fig:comp3}
\end{figure*}

\begin{figure*}
    \centering
    \includegraphics[width=0.95\linewidth]{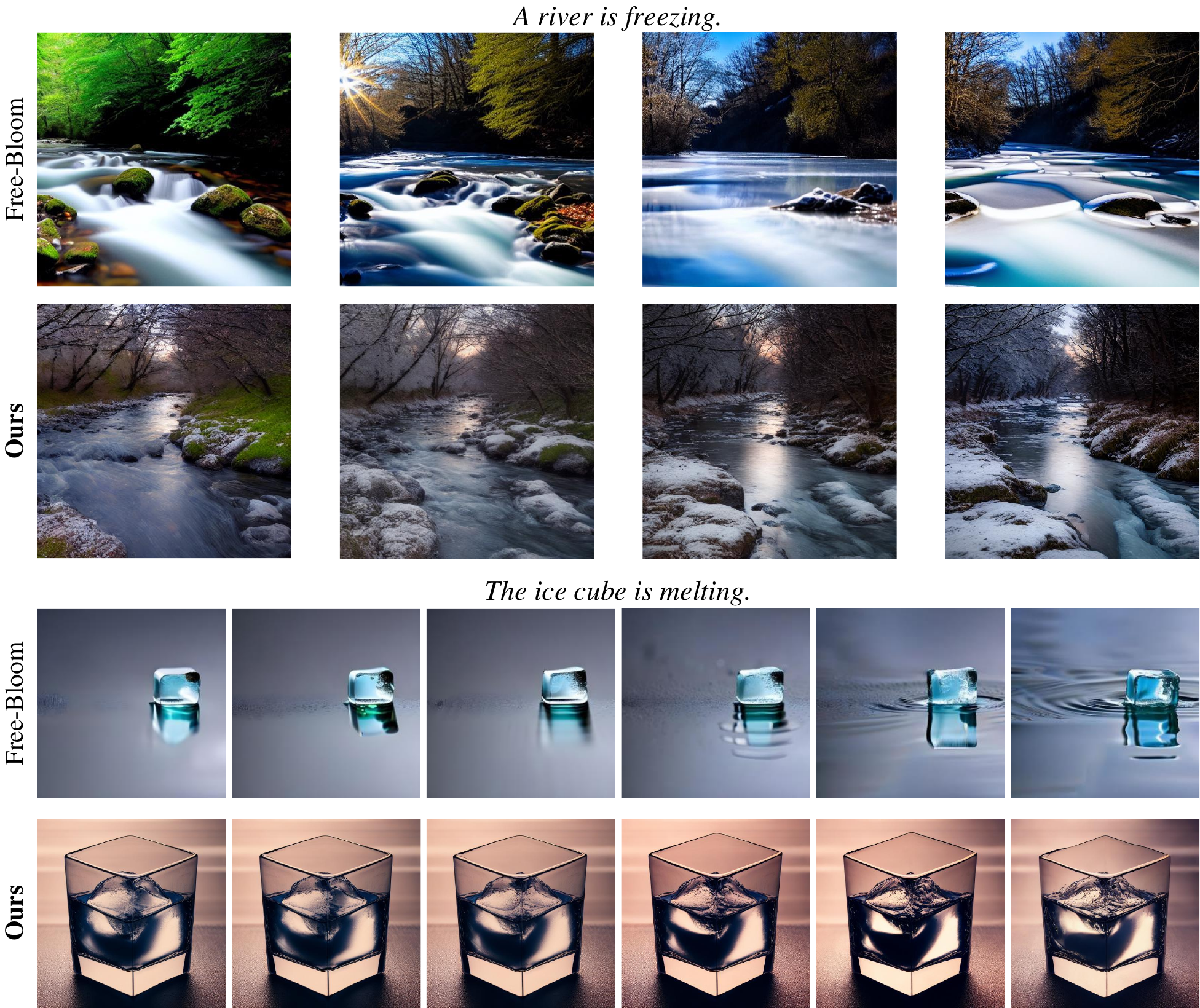}
    \caption{\textbf{Additional comparison visual results with Free-Bloom~\cite{freebloom}.} Ours could also model developing events with better video consistency.}
    \label{fig:comp4}
\end{figure*}

Due to the absence of public code of LVD~\cite{lian2023llmgroundedvideo} and Free-Bloom~\cite{freebloom}, we draw examples from their original papers for further comparison. As shown in Fig.~\ref{fig:comp3} for comparison with LVD, though ours do not use pretrained large-scale video models as base models, ours still achieve concise location control. Moreover, ours are better in object appearance preservation indicated by the ``rock". And we could also model the camera movement as shown in the example of car.
Besides examples modelling object movement, more examples modelling events development like ``freezing" are provided for comparison with Free-Bloom in Fig.~\ref{fig:comp4}. As observed, ours not only excel in replacement control but also are competitive in event development control.

\begin{figure*}
    \centering
    \includegraphics[width=1\linewidth]{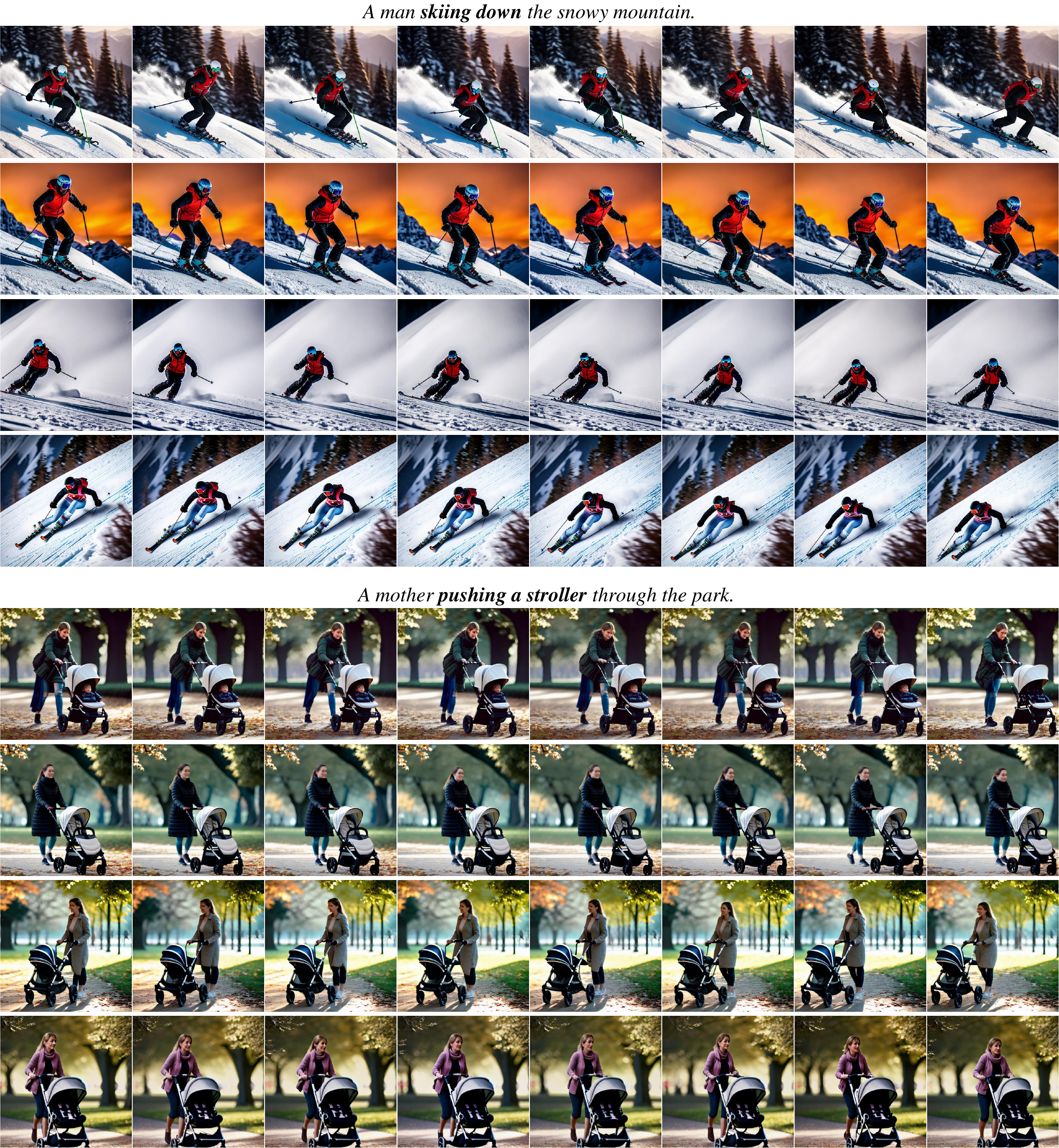}
    \caption{\textbf{Different samples of the same prompt using our method.}}
    \label{fig:more}
\end{figure*}

\begin{figure*}
    \centering
    \includegraphics[width=1\linewidth]{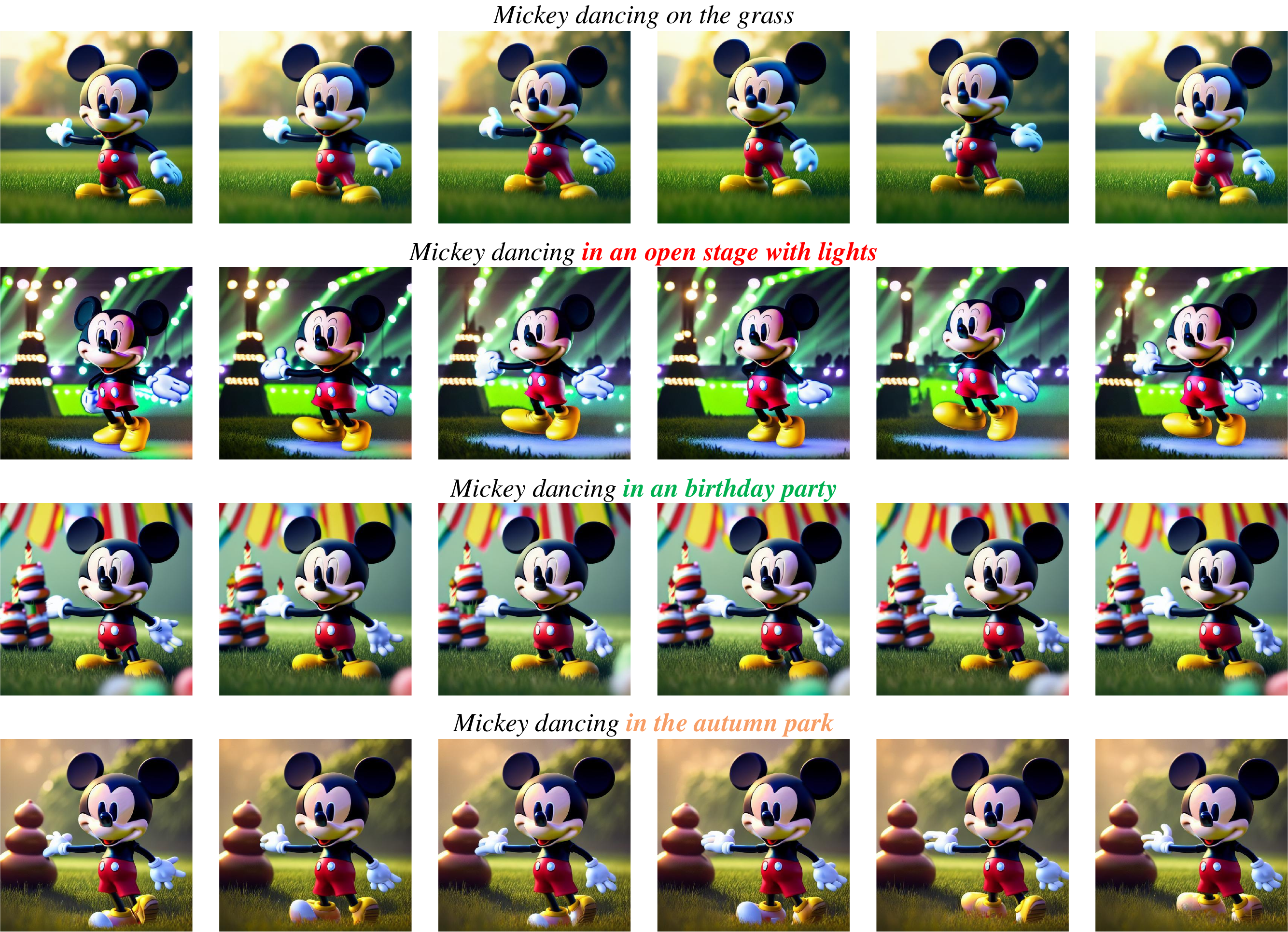}
    \caption{\textbf{Background edit while preserving the foreground.} As seen, our disentangled design allows for editing background while preserving the semantics of the foreground with reasonable harmonization.}
    \label{fig:edit_background}
\end{figure*}

\begin{figure*}
    \centering
    \includegraphics[width=0.9\linewidth]{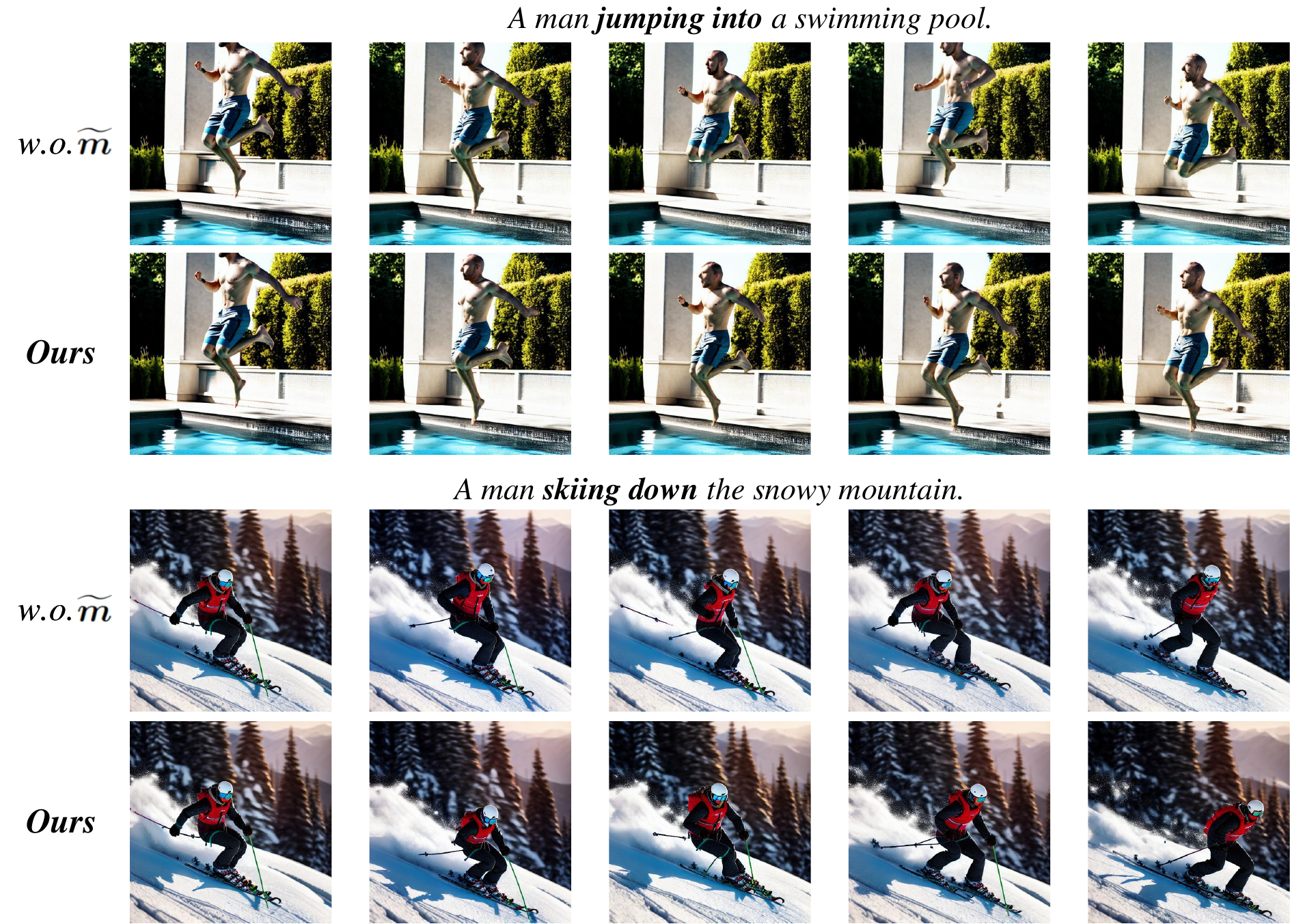}
    \caption{\textbf{Ablation study on the effect of $\widetilde{m}$}}
    \label{fig:wom}
\end{figure*}

\begin{figure*}
    \centering
    \includegraphics[width=1\linewidth]{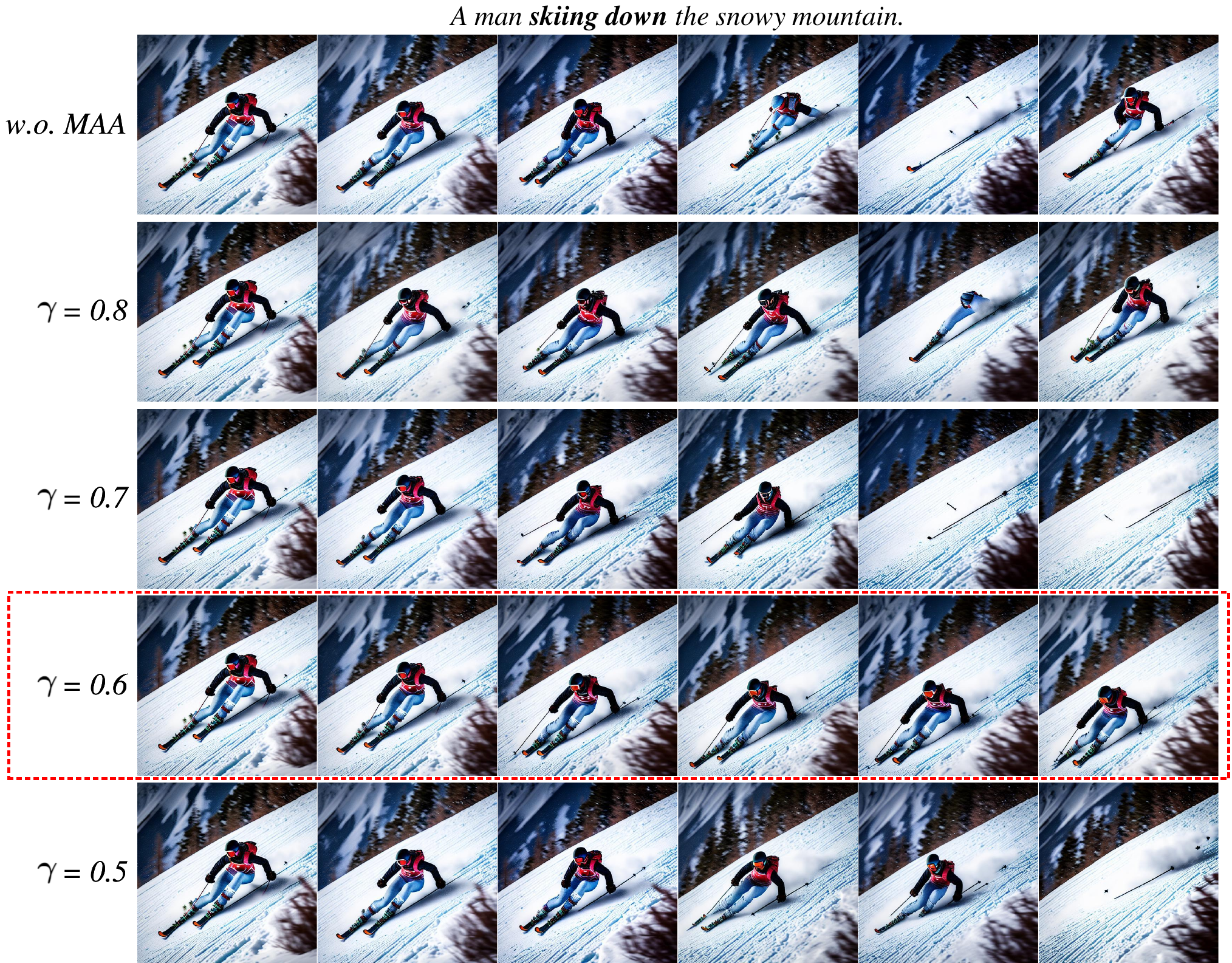}
    \caption{\textbf{Ablation study on different $\gamma$}}
    \label{fig:gamma}
\end{figure*}

\subsection{Different Visual Samples}
In Fig.~\ref{fig:more}, we demonstrate different video samples given the same prompt. As seen, our generation is robust to different seeds. Moreover, for motion without specific directions like ``pushing a stroller", our strategy that inferring extra motion priors from the first frame contributes to synthesizing reasonable videos.

\subsection{Edit Background and Foreground}
As shown in Fig.~\ref{fig:edit_background}, we edit the semantics of the background while preserving the appearance and layout of the foreground with reasonable harmonization. The results further prove the effectiveness and versatility of our disentangled control in zero-shot video edit.

\section{Additional Ablation Study}
\label{ablation_add}
\subsection{Effect of $\widetilde{m}$}
As described in Disentangled Motion Control, $\widetilde{m}$ serves as covering the moving objects in previous frames with surrounding information like image inpainting. Here we conduct ablation study on the effect of $\widetilde{m}$ by directly removing it as shown in Fig.~\ref{fig:wom}. As observed, moving objects of subsequent frames cannot escape from the original location of the first frame, leading to invalid motion control.

\subsection{Effect of $\gamma$}
More results regarding the effect of $\gamma$ are presented in Fig.~\ref{fig:gamma}. Overly large or small $\gamma$ cannot address the issue of object distortion and missing.
\end{appendices}

\end{document}